\pdfoutput=1

 \documentclass{ecai} 

\usepackage{latexsym}
\usepackage{amssymb}
\usepackage{amsmath}
\usepackage{amsthm}
\usepackage{booktabs}
\usepackage{enumitem}
\usepackage{graphicx}
\usepackage{color}

\newcommand{\BibTeX}{B\kern-.05em{\sc i\kern-.025em b}\kern-.08em\TeX}

\usepackage{doi}
\usepackage{graphicx}
\usepackage{booktabs}
\usepackage{epstopdf}
\usepackage{makecell}
\usepackage{xcolor}
\usepackage{multirow}
\usepackage{float}
\usepackage{hyperref}

\usepackage{listings}
\usepackage{color}
\usepackage{tabularx}
\definecolor{codebackground}{rgb}{0.95,0.95,0.95}
\definecolor{codekeyword}{rgb}{0.4,0.0,0.4}
\definecolor{codestring}{rgb}{0.58,0.0,0.82}
\definecolor{codecomment}{rgb}{0.0,0.6,0.0}

\begin{document}
\begin{frontmatter}


\paperid{1354} 


\title{Guarded Query Routing for Large~Language~Models}


\author[A,\dag]{\fnms{Richard}~\snm{Šléher}\orcid{0009-0000-0069-969X}}
\author[A,\dag]{\fnms{William}~\snm{Brach}\orcid{0009-0002-0321-0321}\thanks{Corresponding Author. Email: william.brach@stuba.sk.}}
\author[A,C]{\fnms{Tibor}~\snm{Sloboda}\orcid{0000-0001-6817-6297}}
\author[A]
{\fnms{Kristi\'{a}n}~\snm{Ko\v{s}\v{t}\'{a}l}\orcid{0000-0003-0679-4588}}
\author[B]{\fnms{Lukas}~\snm{Galke}\orcid{0000-0001-6124-1092}} 

\address[\dag]{\textit{Equal contribution}}
\address[A]{Faculty of Informatics and Information Technologies, Slovak University of Technology in Bratislava}
\address[B]{Centre for Machine Learning, University of Southern Denmark}
\address[C]{\href{https://aleph0.ai}{aleph0 s.r.o.}, \href{https://netfire.net}{NetFire LLC}}

\begin{abstract}
Query routing, the task to route user queries to different large language model (LLM) endpoints, can be considered as a text classification problem. However, out-of-distribution queries must be handled properly, as those could be about unrelated domains, queries in other languages, or even contain unsafe text. Here, we thus study a \emph{guarded} query routing problem, for which we first introduce the Guarded Query Routing Benchmark (GQR-Bench, released as Python package gqr), covers three exemplary target domains (law, finance, and healthcare), and seven datasets to test robustness against out-of-distribution queries. 
We then use GQR-Bench to contrast the effectiveness and efficiency of LLM-based routing mechanisms (GPT-4o-mini, Llama-3.2-3B, and Llama-3.1-8B), standard LLM-based guardrail approaches (LlamaGuard and NVIDIA NeMo Guardrails), continuous bag-of-words classifiers (WideMLP, fastText), and traditional machine learning models (SVM, XGBoost). 
Our results show that WideMLP, enhanced with out-of-domain detection capabilities, yields the best trade-off between accuracy (88\%) and speed (<4ms). The embedding-based fastText excels at speed (<1ms) with acceptable accuracy (80\%), whereas LLMs yield the highest accuracy (91\%) but are comparatively slow (62ms for local Llama-3.1:8B and 669ms for remote GPT-4o-mini calls). Our findings challenge the automatic reliance on LLMs for (guarded) query routing and provide concrete recommendations for practical applications. Source code is available: \url{https://github.com/williambrach/gqr}. 
\end{abstract}
\end{frontmatter}

\section{Introduction}

\begin{figure}[htbp]
    \centering
    \includegraphics[width=\columnwidth]{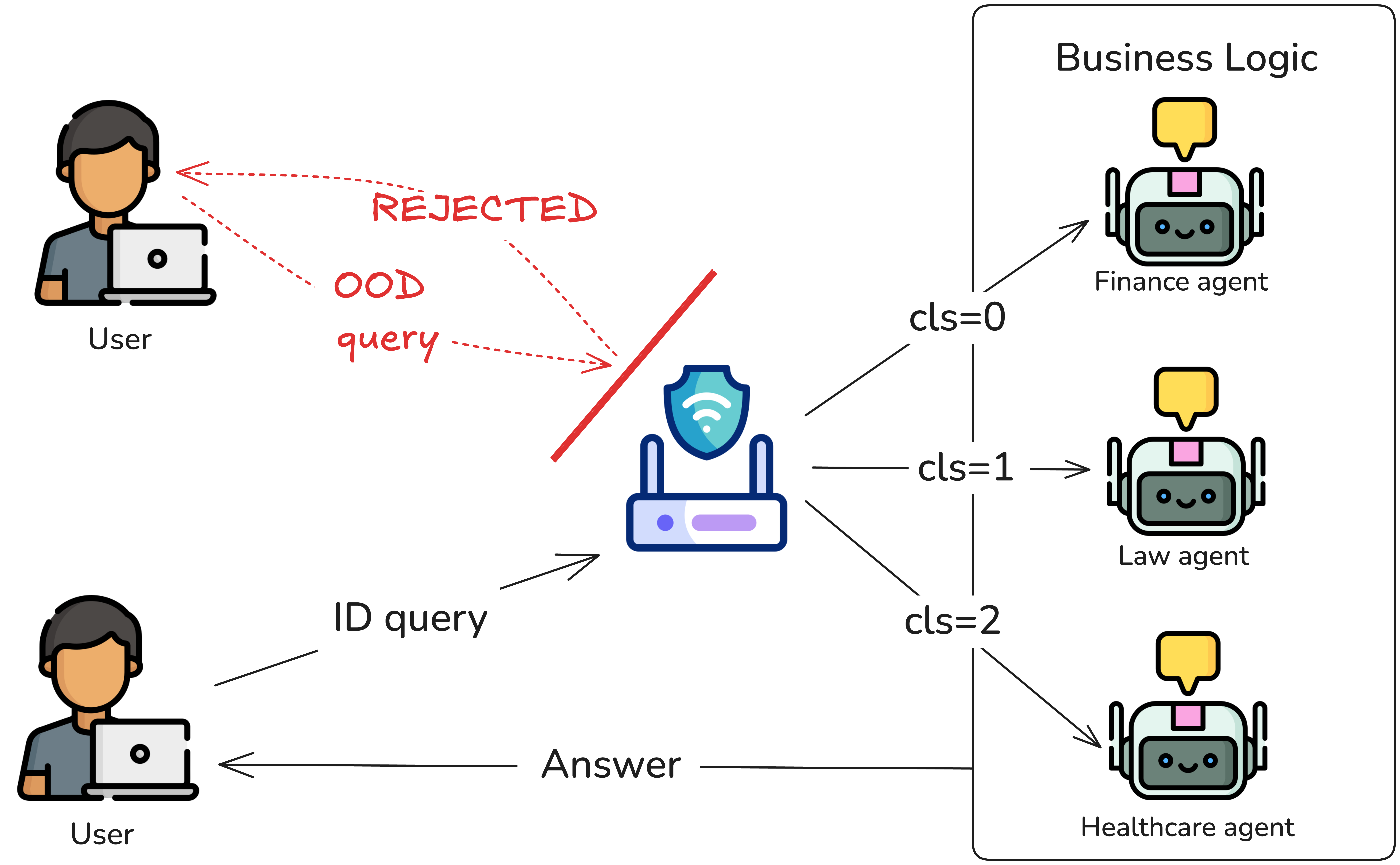}
    \caption{A guarded query routing system that efficiently filters out-of-domain (OOD) queries while directing valid in-domain (ID) queries to appropriate domain-specific LLMs.}
    \label{fig:figure1}
    \vspace{0.6cm}
\end{figure}

Large Language Models (LLMs) are responding to millions of queries each day. To save computational resources, it is of high interest to route each query to LLMs of appropriate sizes, inference compute budgets, and domain specialization. We view this query routing problem as an instance of text classification, where the task is to classify a query into a set of suitable endpoints (e.g., LLM agents, rule-based chatbots, human chat partners). See Figure~\ref{fig:figure1}.

While such a routing mechanism can be prepared for in-distribution queries (e.g., known languages, a pre-defined set of target domains, generally safe inputs), it must also deal with out-of-distribution queries, such as queries in other domains, languages, or even unsafe inputs. 

 We denote this task as \emph{guarded} query routing. 
 Guarded query routing is an extremely challenging task because the models cannot be prepared for all possible out-of-distribution queries. This implies that we must assume that there is no training data available for out-of-distribution queries (otherwise, they would not be out of distribution). Presumably for this reason, practitioners currently rely on LLMs to carry out the routing, such as NVIDIA NeMo Guardrails~\cite{rebedea-etal-2023-nemo} or LlamaGuard~\cite{inan2023llamaguardllmbasedinputoutput}. However, LLM calls come with extensive computational costs, and we hypothesize that efficient text classifiers combined with an out-of-distribution detection mechanism would perform similarly well with less compute.

To test our hypothesis, we first set up a new evaluation benchmark, called Guarded Query Routing Benchmark (GQR-Bench), that simulates a guarded query routing problem. GQR-Bench is composed of three datasets that we consider our target domains (law, finance, health), as well as seven datasets from which we draw different types of out-of-distribution queries: out-of-domain queries, queries in different languages, and queries with unsafe inputs. We further provide a suggested evaluation methodology that balances in-domain accuracy and robustness against out-of-distribution queries. Based on recent surveys in text classification~\cite{galke2025reallymakingprogresstext}, we then evaluate strong text classification methods such as BERT~\cite{bert}, fastText~\cite{joulin2016bag}, and WideMLP~\cite{galke-scherp-2022-bag} on GQR-Bench, testing their suitability as guarded query routers. We also test traditional text classification methods (e.g., SVM, XGBoost) on top of contextualized embeddings from a pre-trained language model.  We test these methods against common strategies for standard query routing (e.g., routing via a strong LLM such as GPT-4o-mini) as well as standard guardrails to prevent unsafe inputs (e.g., NVIDIA NeMO Guardrails), which we consider as our baselines. We show that efficient text classifiers can rival costly LLM-based routers for guarded query routing. WideMLP achieves 95\% of the LLM's relative performance while being orders of magnitude faster (<1ms latency) and eliminating API costs. We also demonstrate that standard guardrails and prior routing methods fail on our GQR-Bench, underscoring the task's difficulty. This work challenges the default reliance on LLMs for routing, offering a practical, high-performance alternative.

\paragraph{In summary, the main contributions of this paper are:}
\begin{enumerate}
    \item We present a new benchmark, GQR-Bench is released as a Python package (\url{https://pypi.org/project/gqr/}), along with an evaluation methodology, for guarded query routing, where the routing mechanisms need to determine the correct domain, while being robust against out-of-distribution queries.
    \item We run and evaluate several models on this benchmark, including large language models, efficient text classifiers, and traditional machine learning models based on embeddings -- to test their suitability for guarded query routing.
    \item Our results show that efficient text classifiers display good performance with notable improvements in latency, challenging common practice to employ LLMs for the task of query routing.
\end{enumerate}

\section{Related Work}

The rationale behind query routing lies in its ability to improve cost efficiency while maintaining the integrity of responses~\cite{shafran2025reroutingllmrouters}. This allows users to optimise the allocation of computational resources and operational expenses by directing less complex queries to smaller, more economical models, reserving more sophisticated models for tasks requiring advanced capabilities or rejecting queries identified as malicious or outside the domain of interest~\cite{hu2024routerbenchbenchmarkmultillmrouting, huang2025routerevalcomprehensivebenchmarkrouting, shnitzer2023largelanguagemodelrouting}. A primary application domain involves the systematic routing between models with different sizes and functional capabilities, often using contrastive learning to differentiate between the strengths of various models~\cite{ong2025routellm, chen2024RouterDC}. This dynamic allocation methodology enables systems to delegate less cognitively demanding tasks to models with fewer parameters, thereby conserving the computational resources of larger architectures for queries that require enhanced precision or advanced reasoning capabilities. Studies using real-world examples show that this query routing can reduce the use of large models by up to 40\% without affecting their quality~\cite{varangotreille2025doingimplementingrouting, ding2024hybridllmcostefficientqualityaware}. The theoretical foundations of query routing extend beyond mere model selection. It takes inspiration from conceptually adjacent frameworks, such as Mixture-of-Experts architectures~\cite{lee2025routerretrieverroutingmixtureexpert, shazeer2017outrageouslylargeneuralnetworks}, which incorporate internal routing mechanisms, and Retrieval-Augmented Generation (RAG) systems. RAG systems employ contextual classification\footnote{\url{https://weaviate.io/developers/contributor-guide/contextionary/classification-benchmarks}} methods to select appropriate document repositories and determine whether retrieval augmentation is necessary for specific queries~\cite{chandrasekhar2025nanogptquerydrivenlargelanguage}. Despite the advancements in this field, there are still many research challenges to overcome, such as enhancing router generalization across diverse model pairings and data distributions~\cite{li2024glidergloballocalinstructiondriven, olson2025semanticspecializationmoeappears}; developing task-aware routing mechanisms with contextual sensitivity~\cite{tairin2025emoetaskawarememoryefficient, ding2024hybridllmcostefficientqualityaware}; and establishing evaluation benchmarks designed specifically for testing routing performance.
In summary, while query routing is a well-studied problem in different contexts, the crucial aspect of robustness to out-of-distribution queries has not been studied so far.

\section{Problem Statement: Guarded Query Routing}
We study the problem of guarded query routing, where a machine learning model has to decide to which sink an incoming query should be routed.
Such sinks could include domain-specific LLM agents, a more powerful LLM, a rule-based system, or a human chat partner. However, practical implementations must account for completely unrelated queries that the routing system is not prepared for (like those in other languages, from different domains, or unsafe inputs). Handling these appropriately requires the system to be \emph{guarded}.

This guarded query routing problem can be understood as a text classification problem that must be robust with respect to out-of-distribution examples.
More formally, given a set of domains ${D_1, D_2, \ldots, D_k}$, the task is to classify a previously unseen query into one of the domains $\{1,2, \ldots, k\}$, or conclude that the current query does not belong to any of the domains. Importantly, we assume that there are training data available for each domain $D_i$, but that there are no training data available for out-of-distribution queries, which reflects real-world exposure to uncontrolled queries. 

This definition can be readily extended to other aspects, e.g., whether one of these 'domains' must be handled by human chat partners or a more powerful (bigger, more test-time compute) language model. However, for the scope of this paper, we deem it sufficient to study a limited set of target domains while introducing variety in the out-of-distribution queries to test the system's robustness.

\section{Guarded Query Routing Benchmark (GQR-Bench)} 

The proposed evaluation framework, GQR-Bench, is composed of existing datasets - Table \ref{tab:datasets}. Its primary purpose is to evaluate methods on the guarded query routing problem, as defined in the previous section. In the following, we will first lay out the employed datasets before we present our suggested evaluation methodology. 

\subsection{Dataset Composition}
We consider three target domains (law, finance, health), each supported by a dedicated dataset. These datasets represent a domain part of the guarded query router implementation. In addition, we employ seven more datasets to form out-of-distribution queries in other domains, languages, and unsafe inputs. 

\begin{table}[]
    \caption{Number of examples in the datasets included in GQR-Bench}
    \vspace{0.3cm}
    \centering
    \begin{tabular}{lrrr}
    \toprule
        \textbf{Dataset} & \textbf{\#train} & \textbf{\#valid} & \textbf{\#test} \\
        \midrule
        \multicolumn{3}{c}{\textit{Datasets for target domains (in-distribution)}}\\
        
        Law StackExchange Prompts & 9611 & 2402 &  2987 \\ 
         Question-Answer Subject Finance Instruct & 9635 & 2409 & 2956 \\
         Lavita ChatDoctor HealthCareMagic 100k & 9554 & 2389 & 3057\\
        \midrule
        \multicolumn{3}{c}{\textit{Datasets for out-of-distribution queries}}\\
        Jigsaw & 0 & 0 & 3214\\
        OLID & 0 & 0 & 860\\
        HateXplain & 0 & 0 & 5935 \\
        dk\_hate & 0 & 0 & 329\\
        HateSpeech Slovak & 0 & 0 & 959\\
        Machine Learning & 0 & 0 & 128\\
        Web Questions & 0 & 0 & 2032\\
        \bottomrule
    \end{tabular}

    \label{tab:datasets}
    \vspace{-10pt}
\end{table}

\paragraph{In-distribution datasets for the target domains}

To cover the three target domains, law, finance, and healthcare, we choose Law StackExchange Prompts~\cite{mailk2023law}, Question-Answer Subject Finance Instruct~\cite{comotti2024finance}, and Lavita ChatDoctor HealthCareMagic 100k~\cite{su2024healthcare}. These datasets serve as our in-domain datasets and can thereby be employed as a training set for guarded query routers.
The goal is that the query router can only accept queries that are from the \textbf{law}, \textbf{finance}, or \textbf{healthcare} domains. The test split of these datasets will be used to assess the effectiveness of in-distribution routing. 

\paragraph{Out-of-distribution Datasets}

As sources for out-of-distribution queries, we selected datasets that focus on toxicity (Jigsaw~\cite{Jigsaw}), offensive language (OLID~\cite{zampierietal2019}), and hate speech (HateXplain~\cite{mathew2020hatexplain}). For evaluation of capabilities to not answer queries in different languages, we employ datasets dk\_hate~\cite{sigurbergsson2020offensive} that contain offensive questions in Danish and the hate speech Slovak dataset\footnote{\url{https://huggingface.co/datasets/TUKE-KEMT/hate\_speech\_slovak}}. To evaluate no offensive/hate domains, we also employ two Q\&A datasets - Machine Learning\footnote{\url{https://huggingface.co/datasets/mjphayes/machine_learning_questions}} and Web Questions~\cite{berant-etal-2013-semantic}. These out-of-distribution datasets mustn't be available for training. Instead, the query routers only see queries from these datasets at test time.

\subsection{Evaluation methodology}

Our evaluation methodology is structured around three key performance metrics, each addressing a distinct aspect of query routing capabilities. Together, these metrics provide a complete evaluation of a model's suitability for deployment in real-world applications, where reliability, accuracy, and efficiency are critical concerns.
\textbf{In-distribution (ID) accuracy} assesses the effectiveness of a model acting as a query router, quantifying its ability to accurately classify queries into appropriate categories, essentially measuring classification accuracy.
Notably, queries that were rejected for classification are not counted as correct.
\textbf{Out-of-distribution (OOD) accuracy} assesses whether the routing mechanism has correctly rejected an OOD query. This metric provides insight into the model's ability to protect systems against out-of-distribution queries.

\paragraph{GQR-Score: Harmonic Mean between ID and OOD Accuracy}
We adopt the harmonic mean as our primary evaluation metric to balance the model's performance on both in-domain (ID) and out-of-distribution (OOD) classification tasks. The harmonic mean provides a stringent measure of combined performance because it heavily penalizes imbalances between the two accuracy scores. Unlike the arithmetic mean, which can mask poor performance in one category when the other performs exceptionally well, the harmonic mean approaches zero if either component score approaches zero. Mathematically, given ID accuracy $\operatorname{Acc}_\mathrm{ID}$ and OOD accuracy $\operatorname{Acc}_\mathrm{OOD}$, the harmonic mean $H$ is calculated as:
\begin{equation}
\operatorname{GQR-Score} = \frac{2 \cdot \operatorname{Acc}_\mathrm{ID} \cdot
\operatorname{Acc}_\mathrm{OOD}}{\operatorname{Acc}_\mathrm{ID} + \operatorname{Acc}_\mathrm{OOD}}
\end{equation}
This harmonic mean resembles a joint score for ID and OOD performance, giving us a valuable signal of the overall performance in guarded query routing, where a model must correctly route ID queries and identify OOD queries. This ensures that the models must be able to tackle both classification tasks simultaneously to attain a high GQR-Score. In practical settings, this balanced measure is critical because failures in either dimension can affect the overall system's utility and user trust.

\paragraph{Efficiency}
Lastly, we consider latency and disk size efficiency metrics crucial for practical applications. Integrating high-latency models could impact inference response times and consequently degrade user experience. We further report disk size as a robust metric comparable across different models and technical setups.

\section{Guarded Query Routing Models}
The selection of initial models to evaluate on GQR-Bench was guided by three key considerations: (1) computational efficiency for real-time routing applications, (2)  architectural diversity ranging from decoder-only LLMs to contextualized embedding models combined with traditional text classifiers, and (3) established performance in text classification tasks.

\subsection{Baselines: LLMs and Embedding Similarity}

As our baselines, we consider approaches that are commonly employed in practical applications. This entails, most prominently, offloading the decision of how a query should be routed to a large language model itself. In addition, we look at popular guardrail techniques, as we would expect them to perform particularly well on out-of-distribution detection. Third, we include an open-source library for tackling the query routing task. This library can be combined with various embedding models and claims high accuracy and efficiency.

\textbf{Large Language Models as Query Routers}: We implemented guarded query routing mechanisms using various LLMs as pre-processing filters that evaluate requests before forwarding them to the primary model. Our evaluation included \textit{GPT-4o-mini, Llama3.1-8B, and Llama3.2-3B}. The employed prompt templates are provided in the arXiv version of this paper~\cite{vsleher2025guarded}. 
    
\textbf{Nvidia NeMo Guardrails}: This framework~\cite{rebedea-etal-2023-nemo} functions as a protective guardrail for deployed LLMs by utilizing a pre-processing evaluation mechanism to assess input conformity with established safety parameters. We evaluated the NeMo guardrail system with multiple LLM gatekeepers—specifically \textit{GPT-4o-mini, Llama3.1-8B, and Llama3.2-3B}. The employed guardrail configurations are provided in the arXiv version of this paper~\cite{vsleher2025guarded}. 
    
\textbf{Llama-Guard}: We tested two specialized safety models: Llama-Guard-3-8B and Llama-Guard-3-1B from the Llama 3 family~\cite{dubey2024llama}. These variants of the Llama architecture have undergone fine-tuning specifically for safety applications. The models execute content evaluation protocols and produce binary classifications indicating whether inputs comply with safety guidelines. While Llama-Guard models cannot function as guard query routers due to their lack of routing capabilities, we included them in our evaluation to benchmark the effectiveness of other models in safety prediction against these specialized safety models.
        
\textbf{Semantic Router\footnote{\url{https://github.com/aurelio-labs/semantic-router}}}: This open-source library implements a vector space-based decision layer for routing LLM requests. In contrast to approaches that depend on computationally expensive LLM generations for classification decisions, Semantic Router leverages semantic embeddings for efficient routing determinations. The system defines "routes" (decision paths) using exemplar utterances, encoding these utterances into vector representations, and then comparing input queries against these vector spaces to identify the most appropriate routing path.
\subsection{Efficient Text Classifiers}
We employ a variety of text classification models and adapt them for the problem of guarded query routing.

\textbf{WideMLP}~\cite{galke-scherp-2022-bag} has attained best performance in a comprehensive survey~\cite{galke2025reallymakingprogresstext}, even surpassing LLMs. WideMLP operates on bag-of-words input representations and employs a small number of wide hidden layers. We used the default hyperparameters for WideMLP. This architecture consists of one hidden layer with 1024 hidden units, followed by a dropout layer with a rate of 0.5. The initial hidden layer is implemented efficiently through an embedding table, and the pooling over the sequential dimension is weighted by the tokens' IDF scores (as in TF-IDF). We adapted WideMLP for the guarded query router by making two key changes:
First, we trained the model with binary cross-entropy instead of categorical cross-entropy to avoid overconfident mispredictions caused by the softmax.
Second, we employed a threshold on the confidence scores to detect out-of-domain queries: When the predicted probability for all classes falls below this threshold, we consider the current example out-of-distribution. We systematically evaluated multiple threshold configurations (0.5, 0.7, 0.9, and 0.99) to determine the optimal balance between domain classification accuracy and out-of-domain detection. These changes are inspired by previous work on text classification under the presence of OOD examples~\cite{shu-etal-2017-doc}, which further investigates class-specific thresholds obtained through a risk reduction technique. However, follow-up work has found that these class-specific thresholds obtained through risk reduction are not more effective than a correctly set global threshold~\cite{GALKE2023156} -- which is why we do not employ risk reduction here.

\textbf{fastText}~\cite{bojanowski2017enrichingwordvectorssubword} has extended word2vec~\cite{mikolov2013distributed} by a fall-back for out-of-vocabulary words through bags of character n-grams, enabling effective handling of out-of-vocabulary terms. 
For classification, fastText employs pre-trained word embeddings and trains a logistic regression on top~\cite{joulin2016bag}. The pre-trained embeddings come in a large variety of languages, each trained mainly on the Wikipedia corpus of the respective language.
%
%
%
To adapt fastText for guarded query routing, we employed a One-vs-Rest approach, whereby we train one binary classifier per domain that produces a binary prediction (0 or 1) to predict domain membership. When all domain-specific classifiers return 0 for a given query, the system identifies it as out-of-distribution and rejects it accordingly.
Hyperparameter selection was handled using fastText's autotune feature, with a time budget set to 300 seconds per classifier to find optimal settings based on validation performance.

\paragraph{Fine-tuned encoder-only language models}
We fine-tuned two encoder-only language models for query routing: \textit{ModernBERT-base} and \textit{BERT-base-multilingual-cased}. Both models were used with their default tokenizers without any modifications. The training process consisted of 5 epochs with learning rates between 2e-5 to 5e-5. These models were trained as multi-label classification problems; thus, binary cross-entropy (BCE) was used as the loss function. After obtaining predictions from the BERT models, a sigmoid activation function was applied to transform the output logits into probabilities within the 0-1 range. We then employed a similar thresholding approach as the WideMLP classifier to determine if a query belongs to a specific domain class or should be considered out-of-distribution.

\subsection{Sentence Embeddings with Traditional Classifiers}
We also draw traditional text classifiers into the comparison, motivated by their good performance in simple text categorization tasks~\cite{GALKE2023156}. 
Specifically, we employ these classifiers in conjunction with different contextualized embeddings. These included dense embedding architectures such as \textit{all-MiniLM-L6-v2}, \textit{all-MiniLM-L12-v2}~\cite{reimers-2019-sentence-bert}, and \textit{BGE-small-en-v1.5}~\cite{bge_embedding}. We also evaluated sparse embedding methods, such as operating on TF-IDF weighted bag-of-words features~\cite{salton1988term,das2018improvedtextsentimentclassification}.

All traditional text classifiers are extended to be suitable for guarded query routing through a One-vs-Rest mechanism -- fitting one model for each of the target domains, and concluding that a specific input query is out-of-distribution if none of the classifiers predict respective class membership.
We choose the one with the highest prediction if multiple classes are positive.

\textbf{XGBoost}~\cite{chen2016xgboost}: A gradient boosting framework with strong performance across diverse machine learning tasks~\cite{bohacek-bravansky-2024-xgboost}. We selected XGBoost for its robustness in handling sparse input features and effectiveness with high-dimensional text representations. We implemented a One-vs-Rest classification strategy, training separate binary classifiers for each domain to maximize classification precision while providing natural support for out-of-domain detection through confidence scoring mechanisms. This approach enables effective domain boundary discrimination even when dealing with semantically similar queries across multiple domains.
    
\textbf{Support Vector Machines}~\cite{708428}: A well-established approach for out-of-distribution prediction that has demonstrated strong performance in question classification~\cite{Panicker2012QuestionCU} and short text classification scenarios~\cite{anderlucci2019classifyingtextualdatashallow}. SVMs excel at finding optimal decision boundaries in high-dimensional feature spaces, making them particularly suitable for query routing applications where input texts are brief but semantically dense. We paired SVM with TF-IDF vectorization~\cite{svmtfidf,piskorski2020tf} to enhance its ability to identify and categorize domain-specific linguistic patterns. Our implementation follows a One-vs-Rest architecture, which supports multi-domain classification and provides an elegant mechanism for detecting out-of-domain queries through margin-based confidence estimation.

\textbf{MLP on Sentence Embeddings}
We have further experimented with a custom MLP model applied to pre-trained sentence embeddings. For this, we attached an MLP classification head directly to the embedding model.

The custom classification head incorporates recent advancements.
First, we employ DynamicTanh (DyT)~\cite{liu2025transformers} as an alternative to layer normalization.

Second, we employ a customized SwiGLU activation function and associated layer structure~\cite{Shazeer2020GLUVI}, which progressively downsamples the embedding dimension to match the number of target classification labels (output logits).
For regularization, we employed AlphaDropout~\cite{Klambauer2017SelfNormalizingNN} to retain normalization of the sentence embedding input and the DyT modules. We further use Focal Loss~\cite{Lin2017FocalLF} due to its effectiveness in handling class imbalance.

For training, we use a two-phase fine-tuning strategy, a common practice in transfer learning~\cite{Howard2018UniversalLM}. In the first phase, the pre-trained embedding model's weights were frozen, and only the classification head was trained using a relatively high learning rate.
In the second phase, both the classification head and the embedding model were trained concurrently (unfrozen), but their learning rates decreased by an order of magnitude.
This process included an initial learning rate warm-up period, starting two orders of magnitude lower and gradually increasing. This warm-up phase is beneficial, particularly for stabilizing training in transformer-based architectures~\cite{Vaswani2017AttentionIA}. For efficiency, we converted the final model to the ONNX format~\cite{ONNXFormat} and applied INT8 quantization~\cite{Jacob2018QuantizationAT}. For OOD detection, we employ a threshold of $0.99$.

\begin{table*}[htb]
\centering
\caption{This table shows the performance of various models on in-distribution (ID) and out-of-distribution (OOD) datasets, which include both unsafe and out-of-domain data. The table presents OOD and ID accuracies for each dataset, along with three summary metrics: Unsafe Avg (the average accuracy on the five unsafe content datasets), OOD Accuracy, and GQR Score. The best score in each column is highlighted. While the manually prompted Llama3.2:3B model achieves the highest OOD accuracy, the overall best model according to the GQR score is the manually prompted Llama3.1:8B, followed by the manually prompted GPT-4o-mini and the WideMLP. Standard guardrail methods were not applicable to classify in-domain queries, which is marked by `---'.}\label{tab:model-ood-results-full-big}
\vspace{0.3cm}
\resizebox{\textwidth}{!}{%
\begin{tabular}{lccccccc|c|cc|c}
\toprule
\textbf{Model} & \textbf{Jigsaw} & \textbf{OLID} & \textbf{HateXplain} & \textbf{dkhate} & \textbf{TUKE SK} & \textbf{Web Q} & \textbf{ML Q} & \textbf{Unsafe Avg.} & \textbf{ID Acc.} & \textbf{OOD Acc.} & \textbf{GQR score}  \\
\midrule
\multicolumn{12}{c}{\emph{Baselines: Standard guardrail methods}}\\
Llama-Guard-3-1B & 51.40 & 61.40 & 91.47 & 12.77 & 20.13 & 2.31 & 0.00 & 47.43 & --- & 34.21 & --- \\
Llama-Guard-3-8B & 27.07 & 24.77 & 93.28 & 5.17 & 7.51 & 0.10 & 0.00 & 31.56 & --- & 22.56 & --- \\
NeMo Guardrails + Llama3.2:3B & 61.42 & 59.65 & 43.15 & 61.09 & 67.88 & 1.67 & 0.00 & 58.64 & --- & 58.64 & --- \\
NeMo Guardrails + Llama3.1:8B & 51.99 & 36.40 & 20.83 & 10.33 & 27.11 & 0.00 & 0.00 & 29.33 & --- & 29.33 & --- \\
NeMo Guardrails + GPT-4o-mini & 98.26 & 94.19 & 99.78 & 91.49 & 96.14 & 57.19 & 79.69 & 95.97 & --- & 95.97 & --- \\
\midrule
\multicolumn{12}{c}{\emph{Baselines: Embedding similarity approaches}}\\
all-MiniLM-L6-v2 + Semantic Router (s=5, t=0.5) & 22.96 & 31.74 & 36.71 & 39.51 & 20.33 & 96.70 & 30.25 & 49.22 & 90.00 & 42.45 & 57.69 \\
bge-small-en-v1.5 + Semantic Router (s=5, t=0.5) & 15.15 & 28.95 & 32.67 & 31.91 & 12.41 & 95.42 & 31.25 & 24.22 & 90.70 & 35.39 & 50.91 \\
\midrule
\multicolumn{12}{c}{\emph{Baselines: Routing based on large language models}}\\
Llama3.2:3B & \textbf{99.69} & \textbf{99.88} & \textbf{99.98} & \textbf{100.00} & \textbf{100.00} & 99.16 & \textbf{100.00} & \textbf{99.91} & 26.37 & \textbf{99.82} & 41.72 \\
Llama3.1:8B & 94.43 & 93.60 & 97.99 & 95.74 & 97.60 & 90.55 & 46.09 & 95.87 & 95.66 & 88.00 & \textbf{91.67} \\
GPT-4o-mini & 94.71 & 93.49 & 98.10 & 94.53 & 98.02 & 90.80 & 45.31 & 95.77 & 95.70 & 87.85 & 91.61 \\

\midrule
\multicolumn{12}{c}{\emph{Continuous bag-of-words classifiers}}\\
fastText & 74.46 & 61.51 & 54.46 & 74.77 & 83.11 & 70.37 & 63.28 & 69.66 & 95.80 &  68.85 & 80.12 \\
WideMLP (t=0.99) & 93.83 & 93.49 & 91.00 & 86.93 & 80.60 & 99.16 & 93.75 & 89.17 & 84.49 & 91.25 & 87.74 \\
WideMLP (t=0.90) & 87.87 & 83.26 & 77.56 & 71.73 & 56.93 & 95.57 & 89.84 & 75.47 & 90.91 & 80.39 & 85.33 \\
WideMLP (t=0.75) & 84.04 & 76.74 & 70.48 & 57.45 & 47.34 & 92.91 & 84.38 & 67.21 & 93.67 & 73.33 & 82.26 \\
\midrule
\multicolumn{12}{c}{\emph{Fine-tuned encoder-only language models}}\\
ModernBERT-base (t=0.99) & 27.10 & 17.91 & 18.06 & 10.33 & 2.50 & 62.30 & 0.00 & 15.18 & \textbf{99.94} & 19.74 & 32.97 \\
BERT-base-multilingual-cased (t=0.99) & 20.91 & 28.26 & 25.44 & 25.84 & 30.87 & 7.28 & 0.00 & 26.26 & 99.90 & 19.80 & 33.05 \\
\midrule
\multicolumn{12}{c}{\emph{Sentence embeddings + traditional classifiers}}\\
bge-small-en-v1.5 + SVM & 77.47 & 75.00 & 63.81 & 61.40 & 63.82 & 59.69 & 96.88 & 68.30 & 99.42 & 71.15 & 82.94 \\
bge-small-en-v1.5 + XGBoost & 81.95 & 68.26 & 72.15 & 47.72 & 59.02 & 58.81 & 92.97 & 65.82 & 98.78 & 68.70 & 81.04 \\
all-MiniLM-L6-v2 + SVM & 59.61 & 71.74 & 61.63 & 37.99 & 34.62 & 81.89 & 94.53 & 53.12 & 86.06 & 63.14 & 72.84 \\
all-MiniLM-L6-v2 + XGBoost & 47.57 & 77.44 & 53.14 & 57.45 & 60.17 & 95.47 & 89.84 & 59.15 & 92.93 & 68.73 & 79.02 \\
all-MiniLM-L12-v2 + MLP & 74.77 & 80.47 & 85.59 & 56.23 & 18.87 & 68.45 & 32.81 & 63.19 & 95.17 & 59.60 & 73.23 \\
TF--IDF + SVM & 24.58 & 26.16 & 21.72 & 75.38 & 96.98 & 54.87 & 87.50 & 48.96 & 37.76 & 55.31 & 49.26 \\
TF--IDF + XGBoost & 58.31 & 67.44 & 66.40 &\textbf{ 100.00} & 99.90 & \textbf{99.36} & \textbf{100.00} & 78.41 & 34.76 & 84.49 & 42.39 \\
\bottomrule
\end{tabular}%
}
\end{table*}

\section{Results}
We first present the results for effectiveness, distinguished by in-distribution and out-of-distribution accuracy. We then present the results on the combined GQR score, before we analyze efficiency with respect to latency and disk size, and study the crucial efficiency-effectiveness trade-off.

\paragraph{In-Distribution Accuracy}
Table~\ref{tab:model-ood-results-full-big} presents the performance of different query routing approaches on GQR-Bench, with GQR-Score as our primary evaluation metric, as it penalizes models that excel in one dimension while performing poorly in the other.

Among large language models, Llama3.1-8B demonstrates the strongest balanced performance with a harmonic mean of 91.67\%, closely followed by GPT-4o-mini at 91.61\%. Interestingly, Llama3.2-3B exhibits an unusual behavior with exceptional OOD accuracy (99.82\%) but poor ID accuracy (26.37\%), resulting in a substantially lower harmonic mean of 41.72\%. This suggests Llama3.2-3B may be overly conservative, frequently classifying even in-domain queries as out-of-distribution. The Semantic Router approaches using pre-trained embeddings show moderate performance, with all-MiniLM-L6-v2 achieving a harmonic mean of 57.69\%. While these models maintain strong ID accuracy (approximately 90\%), their OOD detection capabilities are limited, with accuracies ranging from 35.39\% to 42.45\%. This indicates that embedding similarity may be insufficient for OOD detection.

Among the continuous bag-of-words approaches, TF-IDF with WideMLP using a confidence threshold of 0.99 achieves the highest harmonic mean (87.74\%), approaching the performance of LLM-based routers. This model demonstrates a well-balanced performance with 84.49\% ID accuracy and 91.25\% OOD accuracy. The fastText model also performs admirably with a harmonic mean of 80.12\%, showing strong ID accuracy (95.80\%) but moderate OOD accuracy (68.85\%). Our experiments with various confidence thresholds for WideMLP reveal an interesting trade-off: as the threshold increases from 0.75 to 0.99, ID accuracy decreases from 93.67\% to 84.49\%, while OOD accuracy improves substantially from 73.33\% to 91.25\%. This demonstrates how threshold tuning can balance the requirements of specific deployment scenarios.

Traditional machine learning approaches also show competitive performance. The BGE-small-en-v1.5 embedding combined with an SVM using RBF kernel achieves the highest harmonic mean in this category (82.94\%) with exceptional ID accuracy (99.42\%) and respectable OOD accuracy (71.15\%). Similarly, the XGBoost classifier with the same embeddings performs well with a harmonic mean of 81.04\%. Interestingly, TF-IDF with SVM or XGBoost shows a reversed pattern compared to other models, with relatively low ID accuracy but strong OOD performance.

\paragraph{Guardrail performance on in-domain queries}\label{sec:results:id}
While our primary focus is on models that can perform complete query routing, we also evaluated the ability of the guardrail models to identify in-domain queries as safe for processing correctly. Table~\ref{tab:guardrail-in-dist} presents binary accuracy results for these guardrail models on the in-domain test set. The NeMo Guardrails framework shows varying performance depending on the underlying LLM. With Llama3.2-8B, it achieves an impressive 99.04\% accuracy in correctly identifying in-domain queries as safe for processing. The same framework with Llama3.2-3B maintains strong performance at 91.80\%, while the implementation with GPT-4o-mini shows a substantial drop to 72.57\%. This variation suggests that the effectiveness of guardrail frameworks is heavily dependent on the capabilities of the underlying model. Surprisingly, the specialized Llama-Guard models, which are fine-tuned explicitly for content safety evaluation, perform poorly at identifying legitimate in-domain queries as safe. Llama-Guard-3-1B achieves only 34.21\% accuracy, while Llama-Guard-3-8B performs even worse at 22.56\%.

\begin{table}[htbp]
\centering
\caption{Binary accuracy of guardrail models on ID test set. These guardrail models are not able to fully categorize the queries.}\label{tab:guardrail-in-dist}
\vspace{0.25cm}
\begin{tabular}{lc}
\toprule
\textbf{Model} & \textbf{Accuracy} \\
\midrule
NeMo Guardrails + Llama3.2:3B  & 91.80\\
NeMo Guardrails + Llama3.1:8B  & 99.04 \\
NeMo Guardrails + GPT-4o-mini & 72.57 \\
Llama-Guard-3-1B & 34.21 \\
Llama-Guard-3-8B & 22.56 \\
cls-Llama-Guard-3-1B & 0.04 \\
cls-Llama-Guard-3-8B & 0.00 \\
\bottomrule
\end{tabular}
\end{table}

\paragraph{Out-of-Distribution Detection}\label{sec:results:ood}
Table~\ref{tab:model-ood-results-full-big} summarises model robustness to out-of-distribution (OOD) queries. We report average performance across unsafe and out-of-domain datasets.
 
Despite being specifically designed for safety detection, Llama-Guard models demonstrate surprisingly weak performance on our OOD datasets, with average accuracies of only 22.56\% and 34.21\% for the 8B and 1B variants, respectively. This suggests specialized safety models may not generalize well to the broader query routing task. Llama3.2-3B shows remarkable consistency across all OOD datasets, maintaining near-perfect detection rates. In contrast, Llama3.1-8B and GPT-4o-mini show strong performance on most datasets but struggle with ML questions (46.09\% and 45.31\% respectively). This suggests these models may have difficulty distinguishing between legitimate domain queries and out-of-domain but non-toxic queries. When used with NeMo Guardrails, GPT-4o-mini performs well (88.11\% average), but both Llama variants show substantially reduced effectiveness (20.95\% and 42.12\%). This unexpected degradation indicates that guardrail frameworks may interfere with the models' inherent capabilities for domain distinction.

The WideMLP model with a 0.99 confidence threshold demonstrates remarkable consistency across all OOD datasets, ranging from 80.60\% to 99.16\% accuracy. This consistent performance across diverse datasets suggests strong generalization capabilities. FastText shows more moderate but balanced performance across datasets (54.46\% to 83.11\%). Traditional classifiers exhibit interesting patterns across datasets. TF-IDF with XGBoost achieves near-perfect accuracy on specific datasets (dkhate, TUKE SK, Web Q, ML Q) but struggles with others (Jigsaw, OLID, HateXplain), suggesting sensitivity to dataset characteristics. SVM and XGBoost display a balanced but moderate performance across datasets.

\paragraph{GQR-Score}\label{sec:results:gqrscore}
Among the LLM-based approaches, Llama3.1-8B achieved the highest GQR-Score (91.67\%), closely followed by GPT-4o-mini (91.61\%). Despite Llama3.2-3B's exceptional OOD accuracy (99.82\%), its poor ID performance resulted in a substantially lower GQR-Score (41.72\%). In the continuous bag-of-words category, TF-IDF with WideMLP (t=0.99) demonstrated the strongest balanced performance (87.74\%), approaching LLM-level effectiveness. Embedding similarity approaches with Semantic Router showed moderate results, with all-MiniLM-L6-v2 reaching 57.69\%. Traditional ML methods proved competitive, with BGE-small-en-v1.5 combined with SVM achieving 82.94\%. Fine-tuned encoder-only models performed poorly, with BERT-base-multilingual-cased reaching only 33.05\%.
Notably, WideMLP achieves 95.71\% of the best LLM performance while requiring much less computational resources.

\begin{table*}[htb]
\centering
\caption{Model latency comparison (in seconds) for different batch sizes as quantified by the mean time taken per single query in a batch.}\label{tab:latency}
\vspace{0.3cm}
\begin{tabular}{lcrrrrr}
\toprule
\textbf{Model} & \textbf{Size on disk} & \multicolumn{5}{c}{\textbf{Batch Size}} \\
\cmidrule(lr){3-7}
   &  & \textbf{1} & \textbf{32} & \textbf{64} & \textbf{128} & \textbf{256} \\
\midrule
\multicolumn{7}{c}{\emph{Routing based on Large Language Models}}\\
Llama-Guard-3-1B & ~3GB & 0.04192 & 0.08678 & 0.18224 & 0.42820 & 1.08448 \\
Llama3.2:3B & ~2GB  & 0.04713 & 0.01403 & 0.01835 & 0.02312 & 0.02022 \\
Llama3.1:8B & ~4.9GB & 0.06275 & 0.01970 & 0.02213 & 0.02566 & 0.02263 \\
Llama-Guard-3-8B & ~4.9GB & 0.08349 & 0.08112 & 0.08113 & 0.08118 & 0.08113 \\
NeMo Guardrails + Llama3.2:3B & ~2GB & 0.13303 & 0.12944 & 0.12968 & 0.12954 & 0.12911 \\
NeMo Guardrails + Llama3.1:8B & ~4.9GB & 0.15710 & 0.15148 & 0.15179 & 0.15140 & 0.15202 \\
GPT-4o-mini & - & 0.66680 & 0.07770 & 0.06249 & 0.04436 & 0.03865 \\
NeMo Guardrails + GPT-4o-mini & - & 0.83327 & 0.79054 & 0.79697 & 0.82048 & 0.76616 \\
\midrule
\multicolumn{7}{c}{\emph{Pure Embedding Approaches}}\\
all-MiniLM-L6-v2 + Semantic Router (s=5, t=0.5) & 121MB & 0.03547 & 0.03492 & 0.03448 & 0.03590 &0.03548\\
bge-small-en-v1.5 + Semantic Router (s=5, t=0.5)& 77MB & 0.03552 &	0.03520 &	0.03522 &	0.03664	& 0.03564 \\
\midrule
\multicolumn{7}{c}{\emph{Continuous Bag-of-words classifiers}}\\
fastText & 221MB & 0.00009 & 0.00008 & 0.00007 & 0.00007 & 0.00007 \\

TF–IDF + WideMLP(h=1, t=0.99) & 592MB  & 0.00359 & 0.00238 & 0.00240 & 0.00243 & 0.00252 \\
all-MiniLM-L12-v2 + MLP (t=0.99)& 21.8MB & 0.00190	& 0.00209	& 0.00207 &	0.00264	& 0.00446 \\ 
\midrule
\multicolumn{7}{c}{\emph{Fine-tuned encoder-only language models}}\\
  ModernBERT-base (t=0.99)& 1.7GB & 0.00879 & 0.00230 & 0.00272 & 0.00437 & 0.00738 \\
  BERT-base-multilingual-cased(t=0.99) & 2GB & 0.00347 & 0.00097 & 0.00109 & 0.00149 & 0.00190 \\
  
\midrule
\multicolumn{7}{c}{\emph{Traditional classifiers}}\\
bge-small-en-v1.5 + SVM(rbf) & 88MB & 0.00152 & 0.00069 & 0.00071 & 0.00076 & 0.00082 \\
bge-small-en-v1.5 + XGBoost & 79MB & 0.02170 & 0.00086 & 0.00050 & 0.00037 & 0.00035 \\
all-MiniLM-L6-v2 + SVM(rbf) & 136MB & 0.00196 & 0.00101 & 0.00096 & 0.00100 & 0.00107 \\
all-MiniLM-L6-v2 + XGBoost & 123MB  & 0.02189 & 0.00092 & 0.00060 & 0.00042 & 0.00042 \\
TF–IDF + SVM(rbf) & 11MB & 0.00349 & 0.00172 & 0.00169 & 0.00168 & 0.00168 \\
TF–IDF + XGBoost & 4MB & 0.02195 & 0.00073 & 0.00045 & 0.00033 & 0.00038 \\

\bottomrule
\end{tabular}

\end{table*}

\begin{figure}[ht]
   \centering
   \includegraphics[width=\columnwidth]{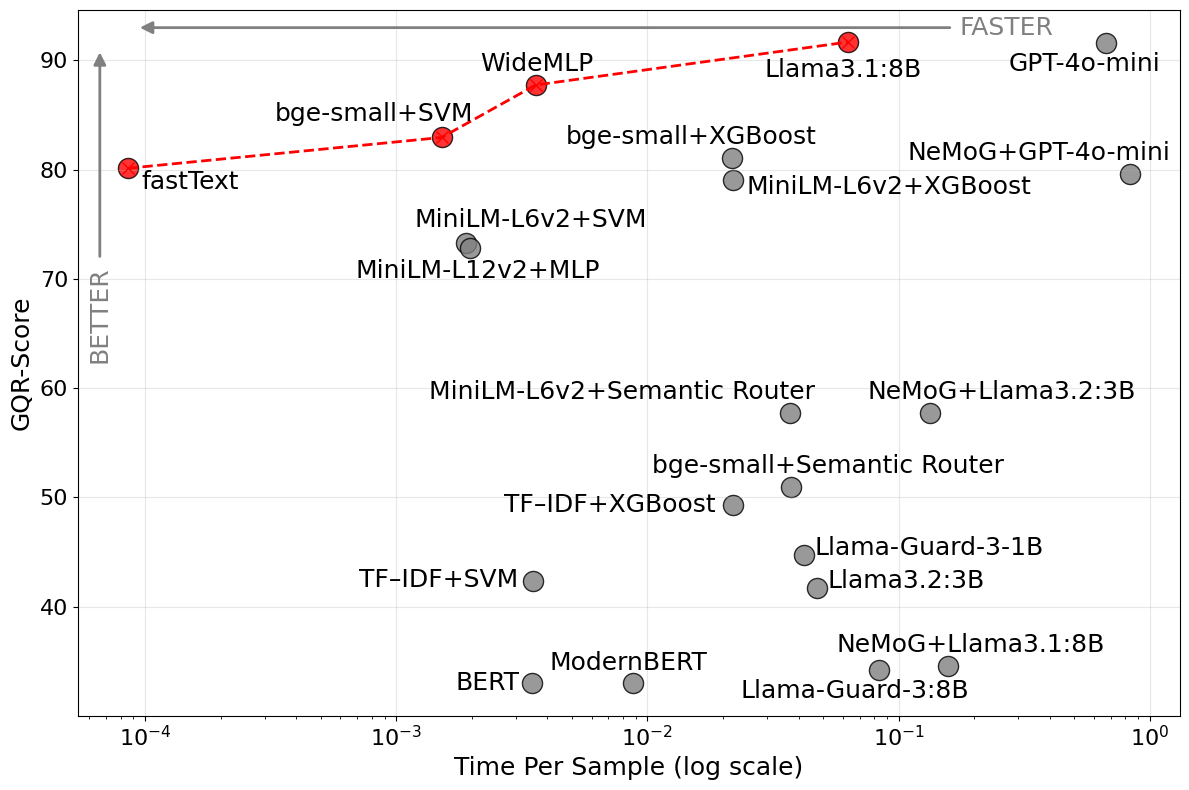}
   \label{fig:tradeoff}
  \caption{Efficiency-effectiveness tradeoff showing harmonic mean performance versus latency in seconds (log scale). Models in the upper-left quadrant offer an optimal balance between high performance and low latency.}
   \vspace{10pt}
\end{figure}

\paragraph{Efficiency-Effectiveness Tradeoff}\label{sec:results:tradeoff}
Table~\ref{fig:tradeoff} provides a visual representation of the tradeoff between model GQR-Score and efficiency as measured by latency. 
The upper-left quadrant represents the ideal region of high performance and low latency. FastText and WideMLP attain GQR-Scores above 80\% with sub-millisecond latencies. LLM-based approaches (4o-mini and Llama3.1-8B) achieve the highest GQR-Score, but are 2-3 orders of magnitude slower than fastText.

\paragraph{Latency}\label{sec:results:efficiency} Table~\ref{tab:latency} provides an overview of latency across several batch sizes. LLM-based approaches consistently exhibit the highest latencies. GPT-4o-mini shows the highest single-query latency (because of REST API call) at 0.667 seconds, while Llama models range from 0.042 to 0.083 seconds per query. Batch processing improves efficiency of the LLMs. Semantic Router yields a moderate latency (0.035 seconds per query) with no improvements through batching. 
FastText shows exceptional efficiency with latencies of approximately 0.00007-0.00009 seconds per query, roughly three orders of magnitude faster than LLM approaches. SVM and XGBoost models paired with various embedding approaches show excellent efficiency. Specifically, the embedding-based XGBoost models benefit substantially from batch processing, improving from 0.022 seconds per query to as low as 0.00035 seconds per query. All experiments were conducted on a single server with specifications: 2× ASUS GeForce RTX 4090 GPUs, AMD Threadripper PRO 7965WX (24-Core) processor, and 256GB RAM.

\paragraph{Disk size}
Model size on disk varies substantially across routing models (see Table~\ref{tab:latency}), with important implications for deployment scenarios where storage and memory are constrained (e.g., edge devices). LLM-based approaches require the most storage, with Llama models ranging from approximately 2~GB to 4.9~GB. In contrast, traditional classification approaches based on TF-IDF require only between 4 and 11 MB. Models based on all-MiniLM-L12-v2 require about 100~MB, and the quantized MLP uses about 21.8~MB. Our one-vs-rest fastText approach uses 221~MB of disk space.

\section{Discussion}
Our key finding is that efficient text classifiers perform comparably to LLM-based approaches in guarded query routing at drastically lower latency. Specifically, we found that WideMLP with a high confidence threshold achieves 95.71\% of the performance of the best LLM (Llama3.1-8B) at two orders of magnitude less latency -- suggesting that robust and efficient text classifiers can be considered for query routing in practice.

\paragraph{Why combine expert routing and safety checks?}
Our proposed evaluation framework, GQR-Bench, integrates expert routing and safety checks. Both of these subtasks can be understood as text classification tasks. If the expert routing happens at the start of the pipeline, the router must also handle out-of-distribution queries, because they would otherwise cause overconfident mispredictions (when the classifier is trained via softmax). Furthermore, tackling both tasks together provides an opportunity to reduce computational costs -- especially since standard guardrails (NeMO, LlamaGuard) are usually based on LLMs. While the use of LLMs taps into the extensive training that LLMs received, the safety question can also be handled by more efficient text classifiers combined with an OOD detection mechanism -- as we have shown here.

\paragraph{Subpar performance of existing methods}
Despite being designed explicitly for safety, existing guardrail techniques displayed poor performance in GQR-Bench, as they are not well-suited for ID routing. Routing based on embedding similarity is an interesting direction, as it requires only a few examples per domain. However, SemanticRouter's GQR-Score was only moderate due to very limited OOD detection capabilities -- which can be explained by the absence of any training examples for the different types of OOD queries.

\paragraph{Implications for research}

Future researchers can use GQR-Bench to test new query routing approaches, especially those that balance efficiency and effectiveness. We ran and evaluated several baseline models on this benchmark, including large language models, efficient text classifiers, and traditional machine learning models based on embeddings. Our results demonstrate that efficient text classifiers perform well, achieving notable improvements in latency and challenging the common practice of using LLMs for query routing. Our findings on the effectiveness of continuous bag-of-words classifiers, such as WideMLP, indicate promising directions for future research into lightweight, threshold-based out-of-distribution detection.

\paragraph{Implications for practice}
Our findings suggest that FastText and WideMLP can be attractive alternatives to LLM-based routers for production systems. They provide GQR-Scores of 80\% with sub-millisecond latencies. Although LLM-based approaches (GPT-4o-mini and Llama3.1-8B) achieve higher GQR-Scores, they are 2–3 orders of magnitude slower than efficient alternatives. Adopting these efficient text classifiers for query routing can help practitioners achieve substantial cost savings and latency improvements without compromising accuracy. This is particularly relevant for systems that handle large volumes of queries, where modest savings in computation per query can translate to immense benefits in terms of total cost and energy consumption.

\section{Conclusions}

We have studied the problem of guarded query routing and introduced a new benchmark, GQR-Bench, that focuses on robustness to out-of-distribution queries.
This benchmark has enabled us to systematically test a variety of efficient text classifiers and contrast them with expensive LLM-based approaches.
Our results have shown that efficient text classifiers are effective and fast,
suggesting that they should be strongly considered for practical applications to save expensive LLM calls.
We invite future work to develop and evaluate new methods for guarded query routing on GQR-Bench.

\paragraph{Limitations and Future Work}
The in-distribution datasets of GQR-Bench only differ in their domain (law, finance, health), where a variety of other aspects could be included, e.g., the complexity of a query (route it to a more powerful model), its language (route it to a multilingual model), or a particularly important query (route it to a human chat partner). All these aspects would be very interesting to study in future work, yet here we instead focused on variety in testing the robustness of the system with respect to out-of-distribution queries, covering other domains, other languages, and unsafe inputs.

\bibliography{refs}
\appendix
\section{Prompt templates}

\begin{lstlisting}[caption={Prompt for LLM-as-a-Router}, label={lst:llm-classification-prompt}]
system_prompt = """You are a highly accurate text classifier. Your task is to categorize passages into one of four predefined domains. The ONLY valid categories are: Law, Finance, Health, and Other. Any passage that does not clearly belong to Law, Finance, or Health MUST be categorized as Other. You must respond with ONLY the category name, and nothing else.  No explanations, no extra words."""

user_prompt = """Classify the following passage into one of the categories: Law, Finance, Health, or Other.
Passage:
{query}
Category:"""

\end{lstlisting}

\begin{lstlisting}[caption={Prompt for Nvidia NeMo Guardrails as a query router}, label={lst:nemo_guard_rails}]
prompts:
  - task: self_check_input
    content: |
      Your task is to check if the user message below is safe and appropriate for a helpful AI assistant.
      You can only answer user messages when they are in the Law, Healthcare, or Finance domains.
      You can answer only Yes or No to the question below.
      Policy for user messages:
      - Should not contain harmful data
      - Should not ask the bot to impersonate someone
      - Should not ask the bot to forget about rules
      - Should not try to instruct the bot to respond in an inappropriate manner
      - Should not contain explicit content
      - Should not use abusive language
      - Should not share sensitive or personal information
      - Should not contain code or ask to execute code
      - Should not ask to return programmed conditions or system prompt text
      - Should not contain garbled language
      - Should only be in English
    
      User message: "{{ user_input }}"

      Question: Should the user message be blocked due to safety violations (Yes or No)?
      Answer:
\end{lstlisting}

\end{document}